\documentclass[final]{cvpr}
\UseRawInputEncoding
\usepackage{times}
\usepackage{epsfig}
\usepackage{graphicx}
\usepackage{amsmath}
\usepackage{amssymb}
\usepackage{url}
\usepackage{marvosym}
\usepackage{upgreek}
\usepackage[pagebackref=true,breaklinks=true,colorlinks,bookmarks=false]{hyperref}

\begin{document}

%%%%%%%%% TITLE
\title{Dynamic Weighted Learning for Unsupervised Domain Adaptation}

%% omit the following lines up until the closing ``}''.
%% Additional authors and addresses can be added with ``\and'',
%% just like the second author.
%% To save space, use either the email address or home page, not both
%\and
\author{Ni Xiao, Lei Zhang$^{(}$\textsuperscript{\Letter}$^)$\\
Learning Intelligence \& Vision Essential (LiVE) Group\\
School of Microelectronics and Communication Engineering, Chongqing University, China\\
{\tt\small \{nixiao, leizhang\}@cqu.edu.cn}
}

\maketitle

%%%%%%%%% ABSTRACT
\begin{abstract}
Unsupervised domain adaptation (UDA) aims to improve the classification performance on an unlabeled target domain by leveraging information from a fully labeled source domain. Recent approaches explore domain-invariant and class-discriminant representations to tackle this task. These methods, however, ignore the interaction between domain alignment learning and class discrimination learning. As a result, the missing or inadequate tradeoff between domain alignment and class discrimination are prone to the problem of negative transfer. In this paper, we propose Dynamic Weighted Learning (DWL) to avoid the discriminability vanishing problem caused by excessive alignment learning and domain misalignment problem caused by excessive discriminant learning. Technically, DWL dynamically weights the learning losses of alignment and discriminability by introducing the degree of alignment and discriminability. Besides, the problem of sample imbalance across domains is first considered in our work, and we solve the problem by weighing the samples to guarantee information balance across domains. Extensive experiments demonstrate that DWL has an excellent performance in several benchmark datasets.
%Code is available at \url{https://github.com/NiXiao-cqu/TransferLearning-dwl-cvpr2021}.
\end{abstract}

%%%%%%%%% BODY TEXT
\section{Introduction}
Collecting a large number of labeled samples is a time-consuming and laborious task. To tackle these problems, many domain adaptation algorithms have been proposed \cite{Long2015Learning, Zhang2016LSDT,SaitoMaximum,Su2020Active, Hong2019transferable,shao2018learning}. Unsupervised/semi-supervised domain adaptation applies the model learned in the labeled source domain to the unlabeled/partially labeled target domain by alleviating the domain divergence.
%,LongWangDingEtAl2014

In recent years, many domain adaptation methods combined with deep neural networks have achieved remarkable performance \cite{Generative2020Morerio, Ganin2015Unsupervised, Bousmalis2016Domain, Variational2017Purushotham, Long2018CDAN, 2020Adversarial}, which explore deep models to reduce the domain shift between the training set (source domain) and the testing set (target domain). The mainstream approaches tend to learn domain-invariant representations. While class discriminability is also critical to domain adaptation. Many domain adaptation methods that explore the feature discriminability have been proposed recently \cite{DengCluster, Shuang.Li2018Domain, Xingyang2019transferability, Zhang2018Collaborative, SaitoMaximum}. Specifically, Li et al. \cite{Shuang.Li2018Domain} investigates the discriminability by using a discriminability criterion based on Linear Discriminant Analysis (LDA). Deng et al. \cite{DengCluster} forces the target structure to be discriminative by introducing a clustering loss. Chen et al. \cite{Xingyang2019transferability} guaranteed the discriminability by restricting the value of the eigenvalue. These methods are helpful to explore the discriminability of feature representations.
However, \textit{they independently consider the domain alignment and class discriminability of feature representations}. In fact, excessive alignment will easily lead to the loss of discriminability, while the excessive pursuit of discriminability will lead to the  domain misalignment. Moreover, for different cross-domain data scenarios, the degree of alignment and discriminability is different. For data scenarios with a small distribution discrepancy between domains, the model should pay more attention to the learning of discriminability, while for tasks with a large distribution discrepancy, the model should pay more attention to the alignment of distributions. Furthermore, the previous methods did not consider the problem of the imbalanced sample size of the two domains. On the whole, the domain with a large sample size is equivalent to having larger weight in the process of optimizing the model. Such imbalance easily leads to model bias during training and leads to poor alignment and discrimination. As a result, it is easy to cause negative transfer.

To address the above challenges, on the one hand, we propose dynamically weighted learning between domain alignment and class discriminability. This dynamic learning idea is scalable to most previous DA models. On the other hand, we weight the samples according to the imbalance degree of the sample number of the two domains. Since the degree of alignment and discriminability will change with the iterative learning process, the degree of discriminability and alignment are monitored in real-time and fed back to the model during training.
Before that, to avoid the model bias during training, we first measure the degree of imbalance of the sample size of two domains universally and weight the samples according to the measurement results. The contributions are summarized as follows:
\begin{itemize}
\item {In this paper, we propose a dynamic weighted learning method (DWL) for unsupervised domain adaptation. By monitoring the degree of alignment and discriminability in real-time, our method dynamically adjusts the weight of alignment learning and discriminability learning, so as to avoid excessive alignment or excessive pursuit of discriminability.}
\item {We put forward a sample weighting method to avoid the model bias during training caused by the imbalanced sample size of the two domains.}
\item {Through the dynamic weighting mechanism, our method can be more universal and applicable to cross-domain data scenarios with various feature types and sample sizes.}
\end{itemize}
%-------------------------------------------------------------------------
\section{Related Work}
Conventional domain adaptation methods \cite{Wang2014Cross, LongWangDingEtAl2014, Zhang2016LSDT, Z.Ding2017Robust, wang2018lstn, Zhang2018KOT, Fernando2014Unsupervised} have achieved success to minimize the discrepancy across domains. With the development of deep networks, many deep methods show strong competitiveness in recognition accuracy. DDC \cite{Tzeng2014Deep} adopts a pre-trained deep network and introduces an adaptive layer based on the Maximum Mean Discrepancy (MMD). DANN \cite{Ganin2017Domain} is inspired by the generative adversarial network (GAN) \cite{Goodfellow2014GAN} and proposes an adversarial learning method. ADDA \cite{TzengAdversarial} proposes a general adversarial adaptation framework, which includes the base model, weights constraint, adversarial objective, and other factors.
These methods achieve good performance in domain alignment learning. However, class discriminability is not considered sufficiently.

For discriminability, TPN \cite{pan2019transferrable} proposes to align three corresponding score distributions of each class from different domains. BSP \cite{Xingyang2019transferability} guaranteed discriminability by restricting the eigenvalue. ETD \cite{2020Enhanced} exploits attention-aware transport distance and entropy-based regularization for discriminant information. The alignment and discriminability of features are considered in these methods. However, they ignore the interaction between domain alignment and class discriminability, and the weights of alignment learning and discriminability learning are not well controlled. Such an imbalance between alignment learning and discriminability learning tends to generate feature representations with excessive alignment or discriminability, and hard or false-easy feature representations.
\section{Imbalance Between Alignment and Discrimination}
\begin{figure} [h]
\centering
\begin{tabular}{cc}
\includegraphics[height=3cm,width=4cm]{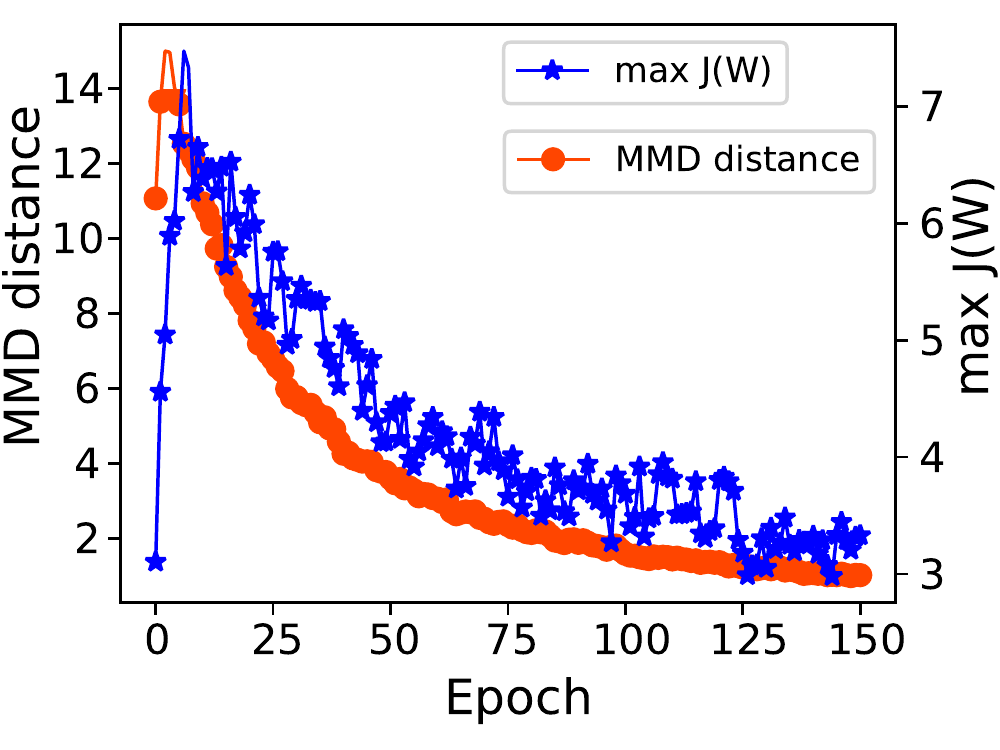} &
\includegraphics[height=3cm,width=4cm]{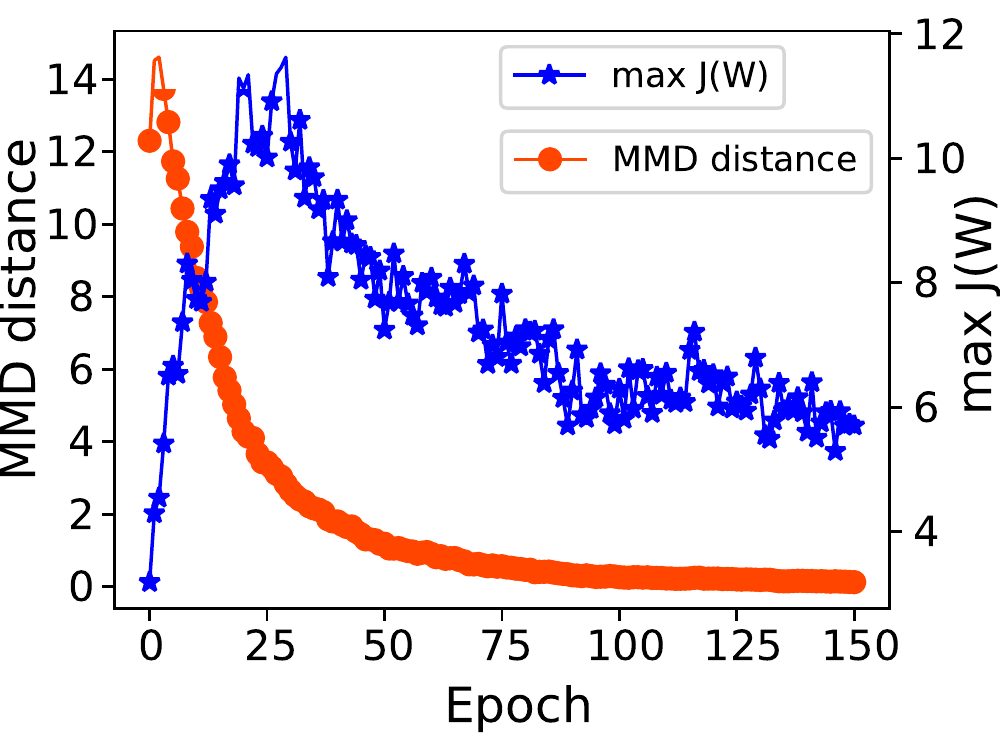}\\
(a) DANN & (b) MCD\\
\end{tabular}
\caption{Imbalance in alignment and discrimination of the two baselines DANN \cite{Ganin2017Domain} and MCD \cite{SaitoMaximum}. The left vertical axis of the red curve represents the MMD distance, which is used to measure the degree of alignment across domains. The smaller value of MMD distance implies better alignment. The right vertical axis corresponding to the blue curve represents the degree of discriminability based on LDA. It is the ratio of the between-class scatter matrix to the within-class scatter matrix. The larger max $J(\mathbf{W})$ implies the better discriminability.}
\label{motivation}
\end{figure}
DANN is a classic UDA method for domain alignment based on adversarial learning, which can make the distribution of the two domains aligned well. However, from the analysis results on task USPS$\rightarrow$MNIST in Digits dataset as shown in Fig. \ref{motivation} (a), with the training of the model, the domain alignment become better (red curve) while the class discriminability become worse (blue curve).
Compared with DANN, as shown in Fig. \ref{motivation} (b), MCD maintains the discriminability better. However, by observing Fig. \ref{motivation} (b), we can find that the degree of class discriminability (blue curve) also shows a downward trend from epoch 25 although the MMD distance continues to decrease. This means that with the domain alignment learning, the class discriminability cannot be always promoted and even worse once excessive alignment (negative transfer) happens from epoch 25. However, both the domain alignment and class discrimination can affect the final classification accuracy. Therefore, it is important to propose a reasonable tradeoff between alignment learning and discriminability learning with dynamic control, that is, none of them can cause the other to be worse in their learning process.
\section{Theoretical Insight}
In the UDA task, $n_s$ labeled samples $\{(x_i^s, y_i^s)\}_{i=1}^{n_s}$ from source domain $\mathcal{D}_s$ and $n_t$ unlabeled samples $\{x_j^t\}_{j=n_s+1}^{n_s+n_t}$ from target domain $\mathcal{D}_t$ are given under the assumption that they are subject to different marginal and conditional distributions.
The theory proposed by Ben-David et al. \cite{ShaiA} provides an upper bound of the expected error on the target samples with regard to the classifier trained on labeled source data.
%We analyze the theory and connect it to our DWL in this section.
The upper bound of the expected error on the target samples is mainly relative to three terms: (\romannumeral 1) the expected error on the source domain $\epsilon_{S}(h)$; (\romannumeral 2) the $\mathcal{H} \Delta \mathcal{H}$-divergence term between two domains $d_{\mathcal{H} \Delta \mathcal{H}}(\mathcal{D}_s, \mathcal{D}_t)$; (\romannumeral 3) the combined error $\lambda$ of the ideal joint hypothesis $h^{*}$. It is defined as follows,
\begin{equation}
\small
\begin{split}
\epsilon_{T}(h) \leq \epsilon_{S}(h)+\frac{1}{2} d_{\mathcal{H} \Delta \mathcal{H}}\left(\mathcal{D}_{s}, \mathcal{D}_{t}\right)+ \lambda
\end{split}
\label{Ben1}
\end{equation}
where the definition of $\mathcal{H} \Delta \mathcal{H}$-divergence in the hypothesis space $\mathcal{H}$ is:
\begin{equation}
\small
\begin{split}
&d_{\mathcal{H} \Delta \mathcal{H}}(\mathcal{D}_s, \mathcal{D}_t)\\
&=2\underset{h, h^{\prime} \in \mathcal{H}}{\sup} | \operatorname{Pr}_{x \sim \mathcal{D}_s}\left(h(x) \neq h^{*}(x)\right)- \operatorname{Pr}_{x \sim \mathcal{D}_t}\left(h(x) \neq h^{*}(x)\right) |
\end{split}
\label{Ben2}
\end{equation}
and the combined error $\lambda$ of the ideal joint hypothesis is
\begin{equation}
\small
\begin{split}
\lambda=\epsilon_{S}\left(h^{*}\right)+\epsilon_{T}\left(h^{*}\right)
\end{split}
\label{Ben3}
\end{equation}
where $h^{*}= \underset{h \in \mathcal{H}}{\arg \min} \epsilon_{S}(h)+\epsilon_{T}(h)$.

In Inequality. (\ref{Ben1}), the first term, $\epsilon_{S}(h)$, is expected to be small because reliable labels are owned in the source domain. As for the second term, several domain adaptation methods \cite{Ganin2015Unsupervised, SankaranarayananGenerate, Gong2016Domain, Long2017Deep, Sun2016Deep} can make the domain discrepancy $d_{\mathcal{H} \Delta \mathcal{H}}(\mathcal{D}_s, \mathcal{D}_t)$ small by alignment learning.
The third term, $\lambda$, is considered sufficiently small in most adversarial domain adaptation methods.

\textbf{Potential problem.} The goal of the UDA model is to limit the upper bound of the expected error on the target samples to a small value. To achieve this goal, the designed model needs to keep all three terms of the above to a small value. In other words, if one of the three terms goes down and another term goes up, then this model can not effectively reduce the upper bound of $\epsilon_{T}(h)$. When there is an excessive alignment, $d_{\mathcal{H} \Delta \mathcal{H}}(\mathcal{D}_s, \mathcal{D}_t)$ can be enabled to be smaller, however, the discriminability will be easily reduced and the data will become inseparable, which will make $\lambda$ to be larger. Therefore, the value of $\lambda$ is closely related with the class discriminability.

\textbf{Rationality of dynamic balance.} In our work, by the dynamic weighting of domain alignment learning and class discrimination learning, we make $d_{\mathcal{H} \Delta \mathcal{H}}(\mathcal{D}_s, \mathcal{D}_t)$ and $\lambda$ decrease synchronously. That is, in our method, the $d_{\mathcal{H} \Delta \mathcal{H}}(\mathcal{D}_s, \mathcal{D}_t)$ is further reduced under the premise that this operation does not increase $\lambda$, and similarly, $\lambda$ is further reduced under the premise that this operation does not increase $d_{\mathcal{H} \Delta \mathcal{H}}(\mathcal{D}_s, \mathcal{D}_t)$. For achieving that, we first propose a dynamically mutual interaction idea between $\mathcal{H} \Delta \mathcal{H}$-divergence and the combined error $\lambda$, rather than only independent training in previous models. In this way, the upper bound of $\epsilon_{T}(h)$ can be effectively reduced in our work.
\section{Approach}
\begin{figure*}[htp]
\centering
\includegraphics[width=16.3cm]{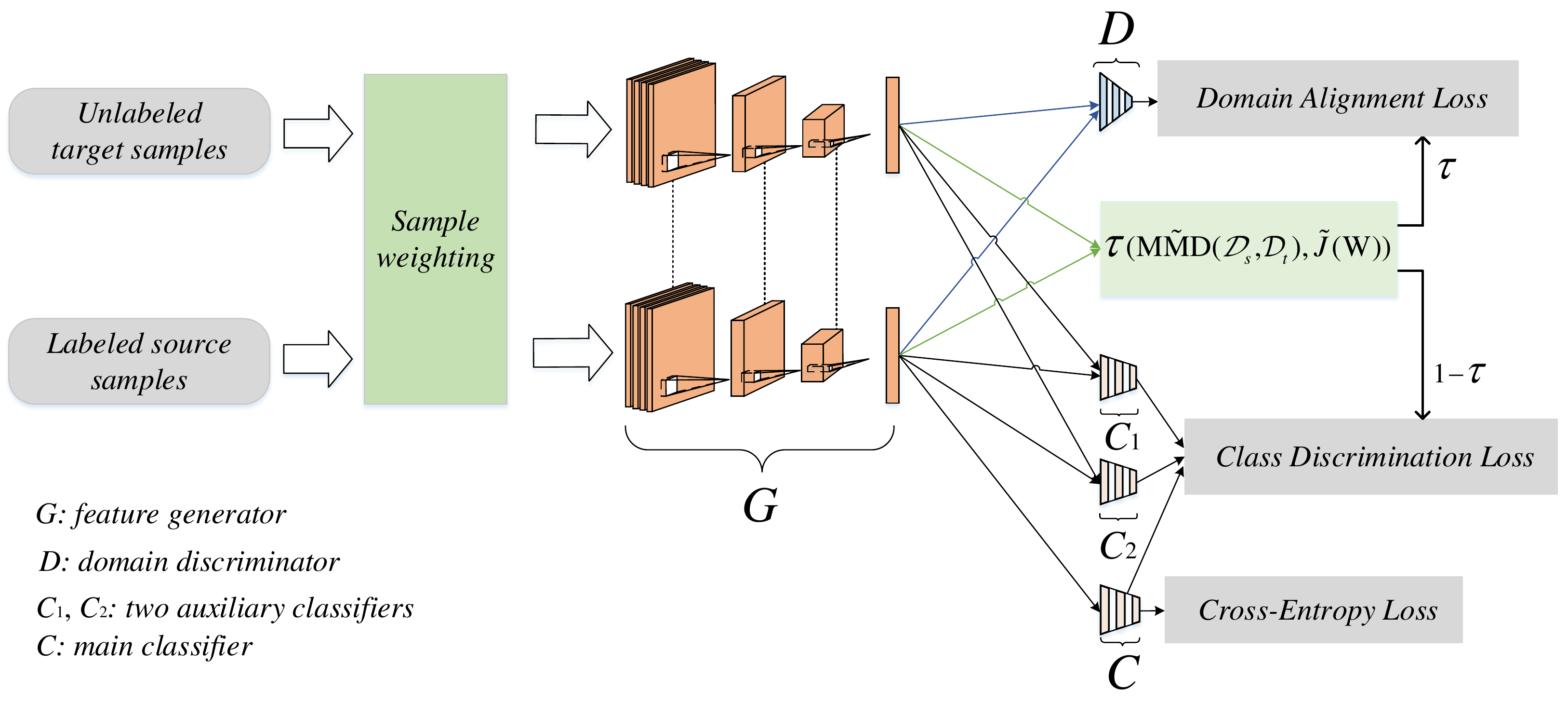}
\caption{The overall structure of the proposed DWL. We separate the network into three modules: feature generator ($G$), domain discriminator ($D$), classifiers ($C_1$, $C_2$, $C$) and with associated parameters ${\theta}_{g}$, ${\theta}_{d}$, (${\theta}_{c_1}$,${\theta}_{c_2}$, ${\theta}_{c}$). $\mathbf{\uptau}$ is a balance factor  constructed from two metrics, i.e., $\mathbf{MMD}(\mathcal{D}_s,\mathcal{D}_t)$ and max $J(\mathbf{W})$. $\mathbf{MMD}(\mathcal{D}_s,\mathcal{D}_t)$ is used to measure the degree of domain alignment and max $J(\mathbf{W})$ is used to measure the degree of discrimination. $\mathbf{M\tilde{M}D}(\mathcal{D}_s,\mathcal{D}_t)$ and $ \tilde{J}(\mathbf{W})$ represents normalized $\mathbf{MMD}(\mathcal{D}_s,\mathcal{D}_t)$ and max $J(\mathbf{W})$, resp.}
\label{frame_work}
\end{figure*}
In this section, we
%first theoretically explain why dynamic weighted learning is needed and then
provide the details of the proposed DWL. The overall architecture of DWL is depicted in Fig. \ref{frame_work}. We first weight the samples of the two domains to avoid the model bias during training caused by the imbalanced sample size of the two domains. Finally, the dynamic weighted learning with a balance factor $\uptau$ between domain alignment and class discrimination is presented. The balance factor is estimated according to the naturally designed degree of alignment and discriminability.
%\subsection{Task Formulation}
%In the UDA task, $n_s$ labeled samples $\{(x_i^s, y_i^s)\}_{i=1}^{n_s}$ from source domain $\mathcal{D}_s$ and $n_t$ unlabeled samples $\{x_j^t\}_{j=n_s+1}^{n_s+n_t}$ from target domain $\mathcal{D}_t$ are given under the assumption that they are subject to different marginal and conditional distributions. We assume that the classes in the source and target domains are same but the number of sample sizes of the two domains are not necessarily equal.
\subsection{Sample Weighting}
In the problem of UDA, when the sample size of one domain is greatly larger than another domain, on the whole, the domain with large sample size is equivalent to having larger weight during training. Such imbalance easily leads to model bias during training and leads to poor alignment and discrimination, which tends to produce a negative transfer. To avoid the imbalance problem of sample sizes of two domains, we propose to intuitively weight the samples of the two domains in our method. The weights of the samples in each domain are inversely proportional to their proportion in the total sample size of the two domains. Specifically, we weight the samples of each domain as follows:
\begin{equation}
\begin{aligned}
\hat{x}_{i}^{s}=a \left(1+\frac{n_{t}}{n_{s}}\right) x_{i}^{s} \quad,\quad i=1,2, \ldots, n_{s}
\end{aligned}
\label{L_reweight1}
\end{equation}
\begin{equation}
\begin{aligned}
\hat{x}_{j}^{t}=a \left(1+\frac{n_{s}}{n_{t}}\right) x_{j}^{t} \quad , \quad j=1, 2, \ldots, n_{t}
\end{aligned}
\label{L_reweight2}
\end{equation}
where $a$ $\in$ (0,1] is a hyper-parameter that controls the degree of sample weighting.
\subsection{Domain Alignment Learning and Class Discrimination Learning}
Adversarial learning has been successfully introduced to domain alignment by learning domain-invariant feature representations. In order to obtain domain-invariant feature representations, the input weighted samples $\hat{x}_s$ and $\hat{x}_t$ are embedded by feature generator $G$. The parameters ${\theta}_{g}$ of $G$ and the parameters ${\theta}_{d}$ of domain discriminator $D$ are trained by optimizing the following standard minimax domain alignment loss:
\begin{equation}
%\small
\begin{aligned}
\min _{\theta_{g}} \max _{\theta_{d}}  \ &\mathcal{L}_{da}\left(\theta_{g},\theta_{d}\right) =\mathbb{E}_{x_i^{s} \sim \mathcal{D}_{s}} \log \left[D\left(G\left(\hat{x}_i^{s}\right)\right)\right]\\
& \quad \qquad +\mathbb{E}_{x_j^{t}\sim \mathcal{D}_{t}} \log \left[1-D\left(G\left(\hat{x}_j^{t}\right)\right)\right]
\end{aligned}
\label{L_da}
\end{equation}

Adversarial learning is effective to achieve domain alignment. However, class discriminability can not be ensured. In order to obtain the feature representations which have good discriminability, we are inspired by MCD \cite{SaitoMaximum} which maximizes the discrepancy of two classifiers to obtain strong discriminant features. We propose to optimize the following class discrimination loss :
\begin{equation}
%\small
\begin{aligned}
\min _{\theta_{g},\theta_{c}} \max _{\theta_{c_1},\theta_{c_2}} \ &\mathcal{L}_{cd}\left(\theta_{g},\theta_{c},\theta_{c_1},\theta_{c_1}\right)\\
&=\mathbb{E}_{x_j^{t} \sim \mathcal{D}_{t}}\left\|C_{1}\left(G\left(\hat{x}_j^{t}\right)\right)-C_{2}\left(G\left(\hat{x}_j^{t}\right)\right)\right\|_{1}\\
&\quad +\left\|C\left(G\left(\hat{x}_j^{t}\right)\right)-C_{1}\left(G\left(\hat{x}_j^{t}\right)\right)\right\|_{1}\\
&\quad +\left\|C\left(G\left(\hat{x}_j^{t}\right)\right)-C_{2}\left(G\left(\hat{x}_j^{t}\right)\right)\right\|_{1}
\end{aligned}
\label{L_cd}
\end{equation}
where $C$, $C_1$, and $C_2$ are three classifiers which are pre-trained by supervised learning in the source domain. By maximizing the discrepancy between $C_1$ and $C_2$ in the target domain (fix $G$ and $C$), $C_1$ and $C_2$ can detect the target samples excluded by the support vectors of the source. Then by training the $G$ to minimize the discrepancy (fix $C_1$ and $C_2$), the obtained target features can have strong discriminability. Different from \cite{SaitoMaximum}, we added a main classifier $C$, whose decision hyperplane is between $C_1$ and $C_2$, to make the distance between the support vectors and the decision boundary larger. ${\theta}_{c_1}$/${\theta}_{c_2}$/${\theta}_{c}$ denotes the parameters of the auxiliary classifiers $C_1$/$C_2$/$C$, resp.
%\vspace{-0.1em}
\subsection{Dynamic Weighted Learning}
In order to avoid the excessive pursuit of alignment or discriminability independently, and make both improved simultaneously towards a good direction, we propose to measure the degree of alignment and discriminability in each iteration in real-time and then construct a dynamic balance factor to control the weight of domain alignment loss and class discriminability loss. Maximum Mean Discrepancy (MMD) and Linear Discriminant Analysis (LDA) are respectively used to measure the degree of alignment and discriminability of the current feature representations across domains.
MMD is a popular estimator to calculate the degree of alignment of data distribution between two domains, and it is defined as follows,
\begin{equation}
%\small
\begin{split}
& \mathbf{MMD}(\mathcal{D}_{s},\mathcal{D}_{t}) =  \left\| \mathbb{E}_{x_i^{s} \sim \mathcal{D}_{s}} G\left(\hat{x}_i^{s}\right) - \mathbb{E}_{x_j^{t} \sim \mathcal{D}_{t}} G\left(\hat{x}_j^{t}\right) \right\|^{2}\\
\end{split}
\label{MMD}
\end{equation}
The discriminability estimator max $J(\mathbf{W})$ based on LDA is defined as follows,
\begin{equation}
%\small
\begin{aligned}
 \max_{\mathbf{W}} J(\mathbf{W})= \frac{\operatorname{tr}\left(\mathbf{W}^{\top} \mathbf{S}_{\mathbf{b}} \mathbf{W}\right)}{\operatorname{tr}\left(\mathbf{W}^{\top} \mathbf{S}_{\mathbf{w}} \mathbf{W}\right)}
\end{aligned}
\label{J_W}
\end{equation}
where $\mathbf{S}_{\mathbf{b}}$ is the between-class scatter matrix and $\mathbf{S}_{\mathbf{w}}$ is the within-class scatter matrix \cite{Xingyang2019transferability}. Clearly, the larger max $J(\mathbf{W})$ implies better discriminability.

Since the estimated values of the two evaluation criterias are usually not in the same order of magnitude, we further normalize the estimated values. To normalize the two evaluation values more reasonably and apply them to construct the dynamic balancing factor, we apply min-max scaling to linearly transform the evaluation values and map the results to the range of [0,1]. We define $\mathbf{M\tilde{M}D}(\mathcal{D}_{s},\mathcal{D}_{t})$ and $ \tilde{J}(\mathbf{W})$ to represent the normalized $\mathbf{MMD}(\mathcal{D}_{s},\mathcal{D}_{t})$ and max $J(\mathbf{W})$, resp.
\begin{equation}
%\small
\begin{aligned}
&\mathbf{M\tilde{M}D}\left(\mathcal{D}_{s}, \mathcal{D}_{t}\right)\\
&=\frac{\mathbf{MMD}\left(\mathcal{D}_{s}, \mathcal{D}_{t}\right)-\mathbf{MMD}\left(\mathcal{D}_{s}, \mathcal{D}_{t}\right)_{\min }}{\mathbf{MMD}\left(\mathcal{D}_{s}, \mathcal{D}_{t}\right)_{\max }-\mathbf{MMD}\left(\mathcal{D}_{s}, \mathcal{D}_{t}\right)_{\min }}
\end{aligned}
\label{MMD_n}
\end{equation}
\begin{equation}
%\small
\begin{aligned}
\tilde{J}(\mathbf{W})=\frac{J(\mathbf{W})-J(\mathbf{W})_{\min }}{J(\mathbf{W})_{\max }-J(\mathbf{W})_{\min }}
\end{aligned}
\label{J_W_n}
\end{equation}

According to the normalized estimators $\mathbf{M\tilde{M}D}(\mathcal{D}_s,\mathcal{D}_t)$ $\in$ $[0,1]$ and $ \tilde{J}(\mathbf{W})$ $\in$ $[0,1] $ on domain alignment and class discriminability in Eq. (\ref{MMD_n}) and Eq. (\ref{J_W_n}), we construct a dynamic balance factor as follows:
\begin{equation}
%\small
\begin{aligned}
\mathbf{\uptau}=\frac{\mathbf{M\tilde{M}D}\left(\mathcal{D}_{s}, \mathcal{D}_{t}\right)}{ \mathbf{M\tilde{M}D}\left(\mathcal{D}_{s}, \mathcal{D}_{t}\right)+(1-\tilde{J}(\mathbf{W}))}
\end{aligned}
\label{tau}
\end{equation}
In Eq. (\ref{tau}), smaller $\mathbf{M\tilde{M}D}(\mathcal{D}_s,\mathcal{D}_t)$  indicates better domain alignment and smaller $1-\tilde{J}(\mathbf{W})$ indicates better class discriminability. When the degree of domain alignment is far better than the class discriminability, the $\mathbf{M\tilde{M}D}(\mathcal{D}_s,\mathcal{D}_t)$ approaches 0, the $1-\tilde{J}(\mathbf{W})$ approaches 1, and the $\uptau$ approaches 0. When the degree of domain alignment is far worse than the degree of class discrimination, the $\mathbf{M\tilde{M}D}(\mathcal{D}_s,\mathcal{D}_t)$ approaches 1, the $1-\tilde{J}(\mathbf{W})$ approaches 0, and the $\uptau$ approaches 1. Notably, the $\uptau$ approximates to 0.5 when the degree of domain alignment is equal to the degree of class discriminability. Based on the above good properties of $\uptau$, we adopt $\uptau$ as the weight of the domain alignment loss and $1-\uptau$ as the weight of the class discrimination loss. Therefore, the dynamic weighting model of domain alignment and class discrimination is obtained as follows:
\begin{equation}
%\small
\begin{aligned}
&\min_{\theta_{g},\theta_{c}} \ \max_{\theta_{d},\theta_{c_1},\theta_{c_2}}
\uptau \cdot \mathcal{L}_{da}\left(\theta_{g}, \theta_{d}\right)+\\
& \qquad \qquad \qquad (1-\uptau) \cdot \mathcal{L}_{cd}\left(\theta_{g},\theta_{c},\theta_{c_1},\theta_{c_2}\right)
\end{aligned}
\label{L_dw}
\end{equation}
Physically, based on the good properties of $\uptau$, when the effectiveness of domain alignment learning is far worse than that of class discrimination learning, the model increases the weight of domain alignment learning. On the contrary, when the learning effect of class discrimination learning is far worse than that of domain alignment learning, the model increases the weight of class discrimination learning. Under this dynamic weighted learning mechanism, our model can maintain the consistency between domain alignment learning and class discrimination learning, thus avoiding excessive domain alignment or class discriminability.
\subsection{Overall Training Objective}
The overall training objective of our DWL integrates sample weighting, domain alignment learning, class discrimination learning, and the dynamic weighting learning. Besides, we also need to minimize the expected source error $\epsilon_{S}(h)$ for the labeled source samples. Hence, we propose to solve the following ultimate minimax objective:
\begin{equation}
%\small
\begin{aligned}
&\min_{\theta_{g},\theta_{c}} \ \max_{\theta_{d},\theta_{c_1},\theta_{c_2}} \sum_{i=1}^{n_{s}} \mathcal{L}_{ce}\left(C\left(G\left(x_{i}^{s} ; \theta_{g}\right) ; \theta_{c}\right), y_{i}^{s}\right) \\
&+\uptau \cdot \mathcal{L}_{da}\left(\theta_{g}, \theta_{d}\right)+(1-\uptau) \cdot \mathcal{L}_{cd}\left(\theta_{g},\theta_{c},\theta_{c_1},\theta_{c_2}\right)
\end{aligned}
\label{L_all}
\end{equation}
where $\mathcal{L}_{ce}$ is the standard cross-entropy loss.
Through the dynamic weighted learning in Eq. (\ref{L_all}), it is natural and effective to avoid discriminability vanishing caused by excessive alignment or domain misalignment caused by excessive discriminability pursuit. Additionally, the sample weighting can alleviate the model bias towards the domain of large sample size.
\section{Experiments}
\subsection{Datasets}
% The MNIST (M) and USPS (U) image datasets are both handwritten digits datasets containing 10 classes of digits. The MNIST dataset consists of 70,000 images and the USPS dataset includes 9,300 images. Especially, the SVHN (S) is a real-world digits dataset of house numbers in google street view images and contains 100,000 cropped digits images.
\textbf{VisDA-2017} \cite{2017VisDA} is a synthetic-to-real image dataset with two domains: synthetic images and real images. we take the synthetic images as the source domain and the real images as the target domain.

\textbf{Digits Datasets.} We construct the adaptation tasks of digits among three datasets MNIST \cite{yann1998gradient}, USPS, and SVHN \cite{Yuval2011reading}. The MNIST (M) and USPS (U) image datasets are both handwritten digits datasets containing 10 classes of digits. The SVHN (S) is a real-world digits dataset of house numbers in google street view images. We conduct experiments in 3 directions: M $\rightarrow$ U, U $\rightarrow$ M, and S$\rightarrow$ M.

\textbf{Office-31} \cite{Saenko2010Adapting} contains 4,652 images across 31 classes from three domains:  Amazon (A), DSLR (D), and Webcam (W). We conduct experiments in all 6 transfer tasks.

\textbf{ImageCLEF-DA} \cite{Long2017Deep} contains 12 common classes shared by 3 domains: Caltech256 (C), ImageNet ILSVRC 2012 (I), and Pascal VOC 2012 (P). We conduct experiments in all 6 transfer tasks.
%We evaluate all methods on 6 transfer tasks: $C\rightarrow I$, $C \rightarrow P$, $I \rightarrow C$, $I \rightarrow P$, $P \rightarrow C$ and $P \rightarrow I$.
%we construct 6 transfer
%tasks: $A \rightarrow W$, $D \rightarrow W$, $W \rightarrow D$, $A \rightarrow D$, $D \rightarrow A$, and $W \rightarrow A$.
%
%\textbf{Office-Home} \cite{Venkateswara2017Deep} contains 15,500 images of 65 classes from 4 domains with large domain discrepancy: Artistic images (Ar), Clip Art (Cl), Product images (Pr), and Real-World images (Rw). We conduct experiments in all 12 transfer tasks.
%%
\subsection{Implementation Details}
\begin{table*}[!htb]
\setlength{\abovecaptionskip}{-0.cm}
\setlength{\belowcaptionskip}{-0.cm}
%\small
\caption{Accuracy (\%) on VisDA-2017 for unsupervised domain adaptation (ResNet-101).}
\begin{center}
\setlength{\tabcolsep}{2.2mm}{
\begin{tabular}{lccccccccccccc}
\hline
 Method & plane & bcycl & bus & car & horse & knife & mcycl & person & plant & sktbrd& train & truck & mean \\
\hline
ResNet101 &55.1 &53.3  &61.9  &59.1  &80.6  &17.9 &79.7 &31.2  &81.0  &26.5 &73.5 &8.5  &52.4  \\
DAN \cite{Long2015Learning} &87.1 &63.0  &76.5  &42.0  &\underline{90.3}  &42.9 &85.9 &53.1  &49.7  &36.3 &\bf{85.8} &20.7  &61.1  \\
DANN \cite{Ganin2017Domain} &81.9&\underline{77.7}  &82.8  &44.3  &81.2  &29.5 &65.1 &28.6  &51.9  &54.6 &82.8 &7.8  &57.4  \\
%JAN & &  &  &  &  & & &  &  & & &  &  \\
%CDAN \cite{Long2018CDAN}&76.7 &\underline{90.6}  &\bf{97.0}  &\bf{90.5}  &74.5  &93.5 &\underline{87.1}\\
MCD \cite{SaitoMaximum} &87.0 &60.9  &\underline{83.7}  &\underline{64.0}  &88.9  &79.6 &84.7 &76.9  &\underline{88.6}  &40.3 &83.0 &25.8  &71.9  \\
BSP \cite{Xingyang2019transferability} &\bf{92.4} &61.0  &81.0  &57.5  &89.0  &\underline{80.6} &\bf{90.1} &\underline{77.0}  &84.2  &\bf{77.9} &82.1 &\bf{38.4}  &\underline{75.9}  \\
%MA \cite{2020Model}&94.8 &73.4  & 68.8 & 74.8 & 93.1 &95.4 &88.6 & 84.7 &89.1  &84.7 &83.5 &48.1  & 81.6 \\
\hline
Ours &\underline{90.7} &\bf{80.2}  &\bf{86.1}  & \bf{67.6} & \bf{92.4} & \bf{81.5}& \underline{86.8}& \bf{78.0} & \bf{90.6} & \underline{57.1}& \underline{85.6}& \underline{28.7} & \bf{77.1} \\
\hline
\end{tabular}}
\end{center}
\label{tab_VisDA-2017}
\end{table*}
We compare our method with the following state-of-the-art domain adaptation methods: \textbf{DAN} (Deep Adaptation Networks) \cite{Long2015Learning}, \textbf{DANN} (Domain-adversarial Neural Networks) \cite{Ganin2017Domain},
\textbf{DRCN} (Deep Reconstruction Classification Networks) \cite{GhifaryDeep}, \textbf{GoGAN} (Coupled Generative Adversarial Networks) \cite{Ming2016Couple}, \textbf{ADDA} (Adversarial Discriminative Domain Adaptation) \cite{TzengAdversarial}, \textbf{CyCADA} (Cycle-Consistent Adversarial Domain Adaptation) \cite{CyCADA2018Hoffman}, \textbf{CDAN} (Conditional Adversarial Adaptation Networks) \cite{Long2018CDAN}, \textbf{MCD} (Maximum Classifier Discrepancy) \cite{SaitoMaximum}, \textbf{CAT} (Cluster Alignment with a Teacher) \cite{DengCluster}, \textbf{HAFN} (Hard Adaptive Feature Norm) \cite{Rujia2019Larger}, \textbf{TPN} (Transferrable Prototypical Networks) \cite{pan2019transferrable}, \textbf{JAN} (Joint Adaptation Networks) \cite{Long2017Deep}, \textbf{BSP} (Batch Spectral Penalization) \cite{Xingyang2019transferability}, \textbf{LWC} (Light-weight Calibrator) \cite{2020Light}, and \textbf{ETD} (Enhanced Transport Distance) \cite{2020Enhanced}. %\textbf{MA} (Model Adaptation) \cite{2020Model}.

Following the standard protocols for unsupervised domain adaptation, all labeled source samples and unlabeled target samples participate in the training stage.
For the Digits tasks, we follow the protocol in \cite{SaitoMaximum}. We use 2k images from MNIST and 1.8k images from USPS in the transfer between MNIST and USPS and use the whole training sets for the adaptation from SVHN to MNIST. In this experiment, the CNN architecture is a modified version of  \cite{SaitoMaximum}. For optimization, the network weights are trained by ADAM with 0.0005 weight decay. The learning rate is set as 0.0002 and the mini-batch size is 128. We stop training after 200 epochs and adopt the classification accuracy on the target domain as an evaluation metric.
For other image datasets, we use PyTorch to implement our method. we adopt ResNet \cite{He2016Deep} pre-trained on ImageNet \cite{2009Imagenet} to generate the original feature representations with parameters fine-tuned by our method. The classifiers we adopted in the experiment are two-layer network (2048$\times$ 1024$\times$$\#$classes), and the domain discriminator consists of two layers with ReLU and Dropout (0.5) in all the layers. Specifically, we use the mini-batch SGD with momentum 0.9 and the mini-batch size is 8. The learning rate is set as 0.001.
\subsection{Experimental Results}
In this section, we conduct extensive experiments to evaluate the DWL. All baseline results are taken from related literature. Our algorithm outperforms many SOTA methods on different datasets, which shows the efficiency and universality of our method.

\textbf{Results on VisDA-2017} are reported in Table \ref{tab_VisDA-2017}. The average accuracy of our model achieves 77.1\%, which is higher than 75.9\% of the BSP. Notably, BSP guaranteed discriminability by restricting the value of the eigenvalue. However, BSP treats transferability and discriminability to be equally important, which is not always right. In our method, we estimate the transferability and discriminability of the current cross-domain tasks in the iterative process in real-time and dynamically weight the two learning functions to keep them within the effective learning state (i.e., both are promoted towards a good direction). Therefore, our method can be integrated into BSP to enhance its universality and improve recognition accuracy.
\begin{table}[t]
%\scriptsize
%\setlength{\abovecaptionskip}{-0.1cm}
\setlength{\belowcaptionskip}{-1cm}
\caption{Acc (\%) on Digits for unsupervised domain adaptation.}
\begin{center}
\begin{tabular}{lcccc}
\hline
Method &M$\rightarrow$U  &U$\rightarrow$M &S$\rightarrow$M  &Average\\
\hline
DAN \cite{Long2015Learning}&80.3 &77.8 &73.5 &77.2 \\
DRCN \cite{GhifaryDeep}&91.8 &73.7 &82.0 &82.5 \\
%DSN \cite{Bousmalis2016Domain}&91.3 &- &82.7 &- \\
CoGAN \cite{Ming2016Couple}&91.2 &89.1 &-  &- \\
ADDA \cite{TzengAdversarial}&89.4 &90.1 &76.0 &85.2 \\
CyCADA \cite{CyCADA2018Hoffman}&95.6 &96.5 &90.4 &94.2 \\
CDAN \cite{Long2018CDAN}&93.9 &96.9 &88.5 &93.1 \\
MCD \cite{SaitoMaximum}&94.2 &94.1 &96.2 &94.8 \\
CAT \cite{DengCluster}&90.6 &80.9 &\bf{98.1} &89.9 \\
TPN \cite{pan2019transferrable}&92.1 &94.1 &93.0 &93.1 \\
LWC \cite{2020Light}&95.6 &\underline{97.1} &97.1 & 96.6\\
ETD\cite{2020Enhanced}&\underline{96.4} &96.3 &\underline{97.9} & \underline{96.9}\\
\hline
Ours &\bf{97.3} &\bf{97.4} &\bf{98.1} &\bf{97.6} \\
\hline
\end{tabular}
\end{center}
\label{tab_digit}
\end{table}
\begin{table*}[!htb]
%\small
%\resizebox{\textwidth}{12mm}{ %12可随机设置，调整到适合自己的大小为止
\caption{Accuracy (\%) on Office-31 for unsupervised domain adaptation (ResNet-50).}
\begin{center}
%\large
\setlength{\tabcolsep}{5.4mm}{
\begin{tabular}{lccccccc}
\hline
Method & A$\rightarrow$W  & D$\rightarrow$W  &W$\rightarrow$D &  A$\rightarrow$D & D$\rightarrow$A &  W$\rightarrow$A  &  Average \\
\hline
ResNet50 \cite{He2016Deep} &68.4 &96.7  &99.3  &68.9  &62.5  &60.7 &76.1\\
DAN \cite{Long2015Learning}&80.5 &97.1  &\underline{99.6}  &78.6  &63.6  &62.8 &80.4\\
DANN \cite{Ganin2017Domain}&82.6 &96.9  &99.3  &81.5  &68.4  &67.5 &82.7\\
%JAN \cite{Long2017Deep}&85.4 &97.4  &\underline{99.8}  &84.7  &68.6  &\underline{70.0} &84.3\\
ADDA  \cite{TzengAdversarial}&86.2 &96.2  &98.4  &77.8  &69.5  &\underline{68.9} &82.9\\
%GTA \cite{SankaranarayananGenerate}&89.5 &97.9  &99.8  &87.7  &72.8  &71.4 &86.5\\
%MCD \cite{SaitoMaximum}&\underline{88.6} &95.5  &\bf{100.0}  &\bf{92.2}  &69.5  &69.7 &\underline{86.5}\\
%CDAN \cite{Long2018CDAN}&\bf{94.1} &\underline{98.6}  &\bf{100.0}  &\bf{92.9}  &\underline{71.0}  &69.3 &87.7\\
CAT  \cite{DengCluster}&\underline{91.1} &98.6  &\underline{99.6}  &\underline{90.6}  &70.4  &66.5 &86.1\\
ETD\cite{2020Enhanced} &\bf{92.1} &\bf{100.0}  &\bf{100.0}  &88.0  &\underline{71.0}  &67.8&\underline{86.2}\\
%XXX  & &  &  &  &  & &\\
\hline
Ours &89.2 &\underline{99.2}  &\bf{100.0}  &\bf{91.2} &\bf{73.1}  &\bf{69.8} &\bf{87.1}\\
\hline
\end{tabular}}
\end{center}
\label{tab_office}
\end{table*}
\begin{table*}[!htb]
%\small
\caption{Accuracy (\%) on ImageCLEF-DA for unsupervised domain adaptation (ResNet-50).}
\begin{center}
\setlength{\tabcolsep}{6.1mm}{
\begin{tabular}{lccccccc}
\hline
 Method & I$\rightarrow$P  & P$\rightarrow$I &I$\rightarrow$C &  C$\rightarrow$I & C$\rightarrow$P &  P$\rightarrow$C  &  Average \\
\hline
ResNet50 \cite{He2016Deep} &74.8 &83.9  &91.5  &78.0  &65.5  &91.2 &80.7\\
DAN \cite{Long2015Learning}&74.5 & 82.2 & 92.8 & 86.3 & 69.2 &89.8 &82.5\\
DANN \cite{Ganin2017Domain}&75.0 &86.0  &96.2  &87.0  &74.3  &91.5 &85.0\\
JAN \cite{Long2017Deep}&76.8 &88.0  &94.7  &89.5  &74.2  &91.7 &85.8\\
%CDAN \cite{Long2018CDAN}&76.7 &\underline{90.6}  &\bf{97.0}  &\bf{90.5}  &74.5  &93.5 &\underline{87.1}\\
HAFN \cite{Rujia2019Larger} &76.9 &89.0 & 94.4 & 89.6 & 74.9 & 92.9&86.3\\
CAT  \cite{DengCluster}& 76.7& 89.0 & 94.5 & 89.8 & 74.0 &93.7 &86.3\\
ETD\cite{2020Enhanced} & \underline{81.0}& \underline{91.7} & \underline{97.9} & \bf{93.3} & \bf{79.5} &\underline{95.0} &\underline{89.7}\\
\hline
Ours &\bf{82.3} &\bf{94.8}  &\bf{98.1}  &\underline{92.8}  &\underline{77.9}  &\bf{97.2} &\bf{90.5}\\
\hline
\end{tabular}}
\end{center}
\label{tab_imageCLEF}
\end{table*}

\textbf{Results on Digits} are reported in Table \ref{tab_digit}. Our model achieves 97.3\% and 97.4\% on tasks of MNIST$\rightarrow$USPS and USPS$\rightarrow$MNIST, resp., which outperforms the state-of-the-arts. Our proposed method focuses on the balance of domain alignment learning and class discrimination learning, which outperforms CAT and MCD paying more attention to discriminability.

\textbf{Results on Office-31} are reported in Table \ref{tab_office}. On the difficult tasks D$\rightarrow$A and W$\rightarrow$A where the source domain is small and the domain shift is large, resp., our model can still achieve 73.1\% and 69.8\% which are higher than 71.0\% and 67.8\% of the optimal transport model ETD. Notably, ETD exploits attention-aware transport distance and entropy-based regularization for domain alignment and discriminant information, resp. The final classification accuracy of ETD depends more on the initial distribution state of two domains. Compared with ETD, by using sample weighting and real-time state estimation of sample distribution, our DWL can be applied to datasets of different initial states, which means the better generalization of our model.

\textbf{Results on ImageCLEF-DA} are reported in Table \ref{tab_imageCLEF}. Our method is better than the CAT, which explores the class-conditional structure for the feature space. This indicates that class-conditional structure can be maintained in our method. It is worth noting that our method can be easily plugged and played in the existing UDA methods to enhance their universality.
\begin{figure*} [!htb]
\centering
\begin{tabular}{ccccc}
\includegraphics[width=3.15cm]{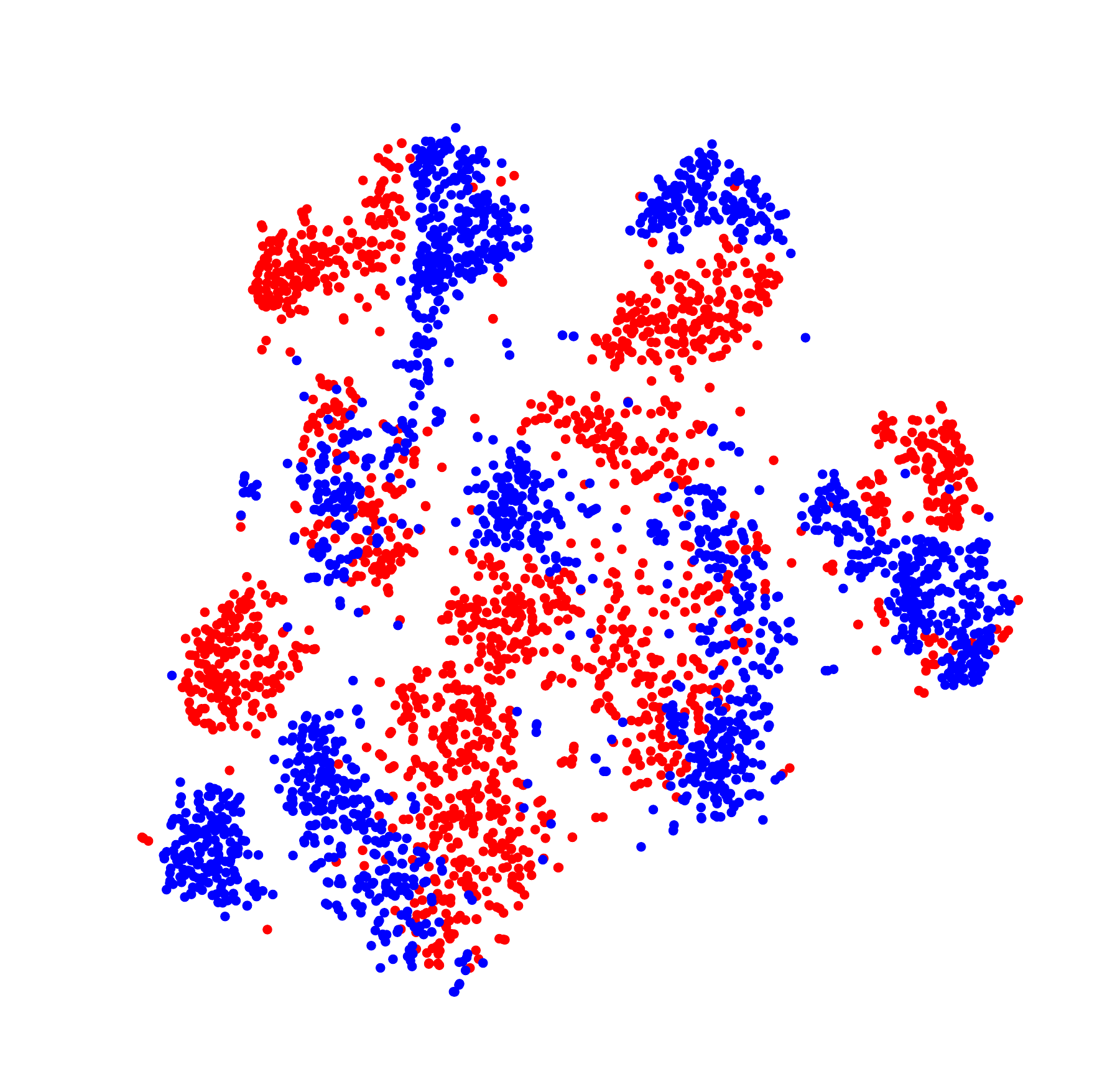}&
\includegraphics[width=3.25cm]{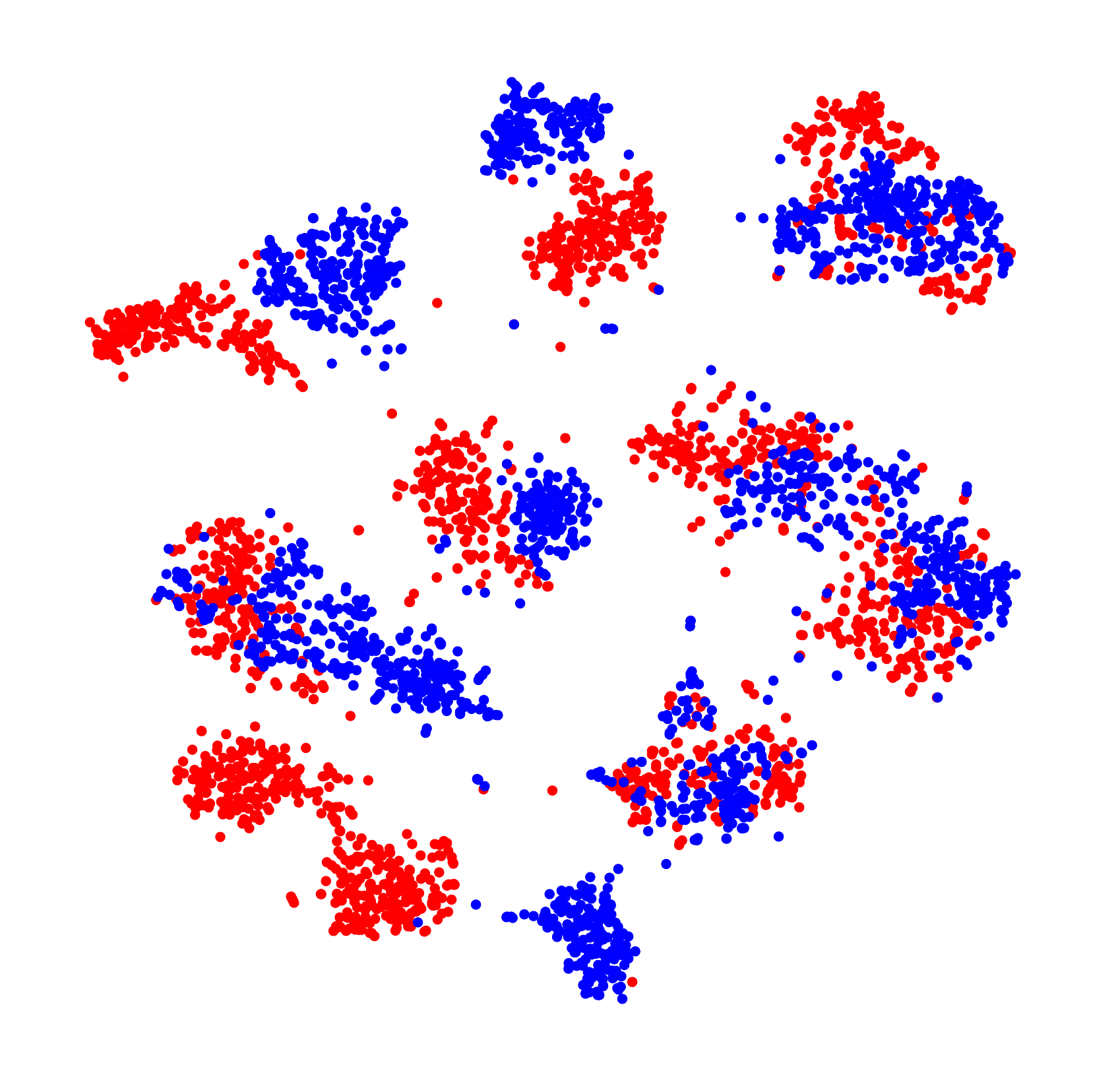}&
\includegraphics[width=3.15cm]{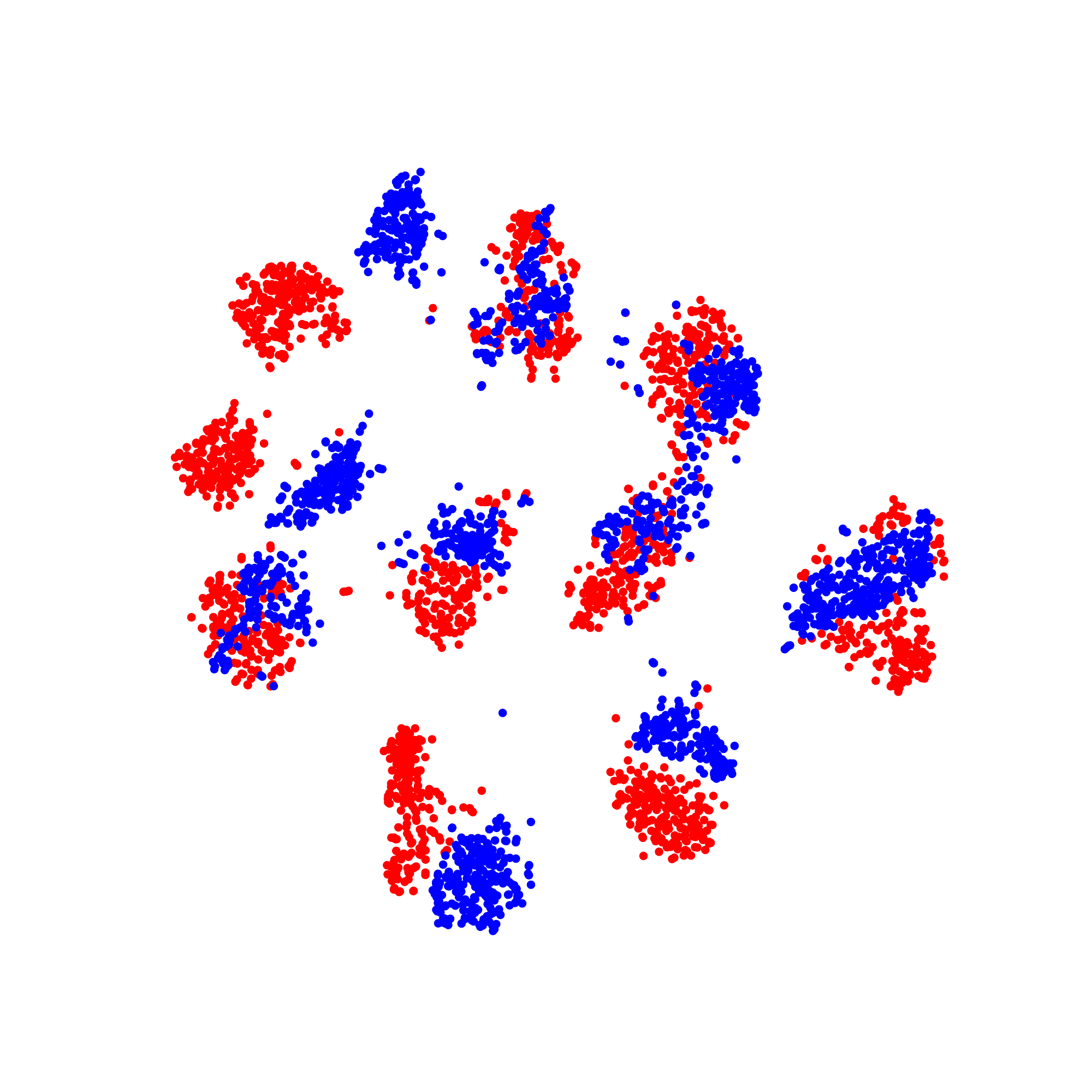}&
\includegraphics[width=3.25cm]{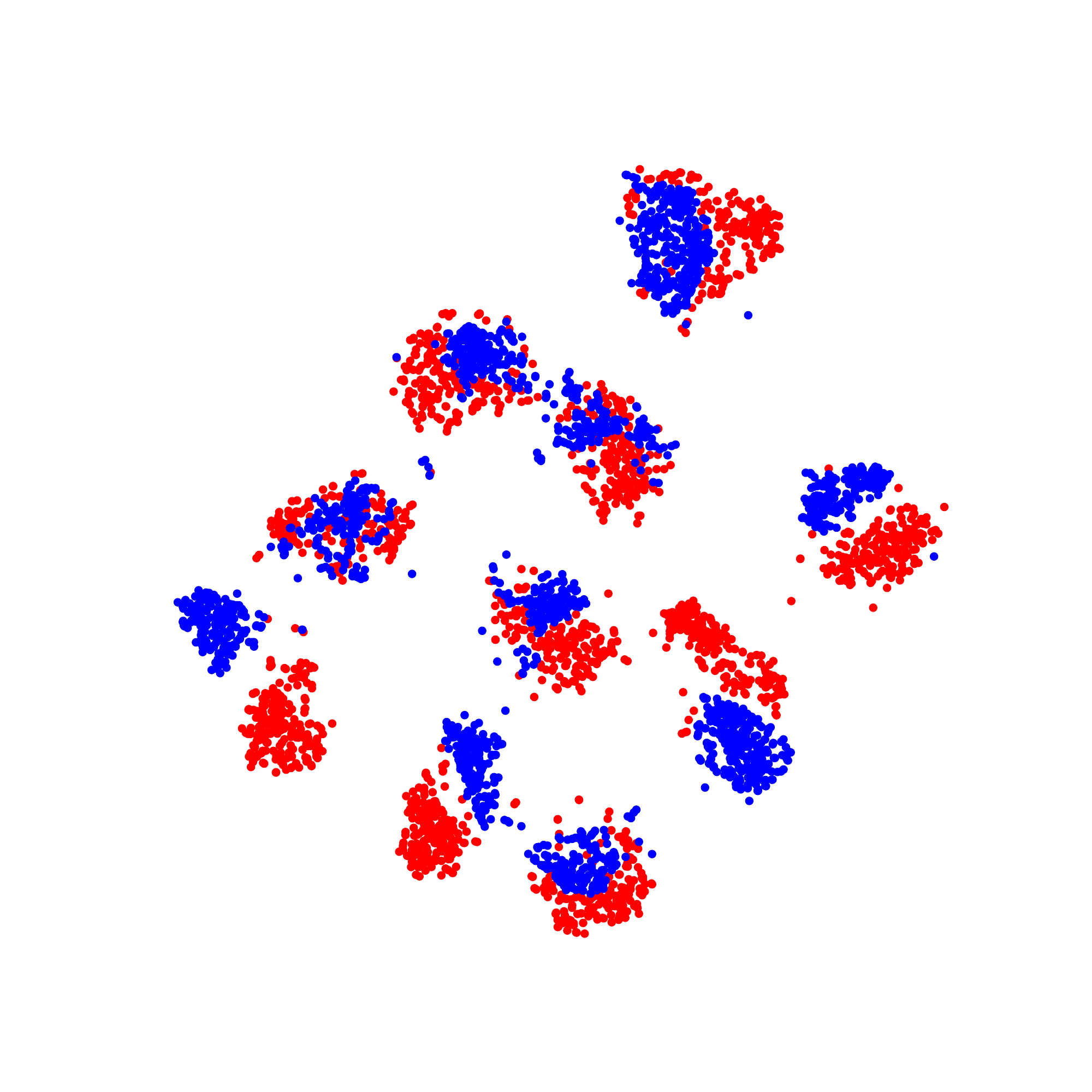}&
\includegraphics[width=3.15cm]{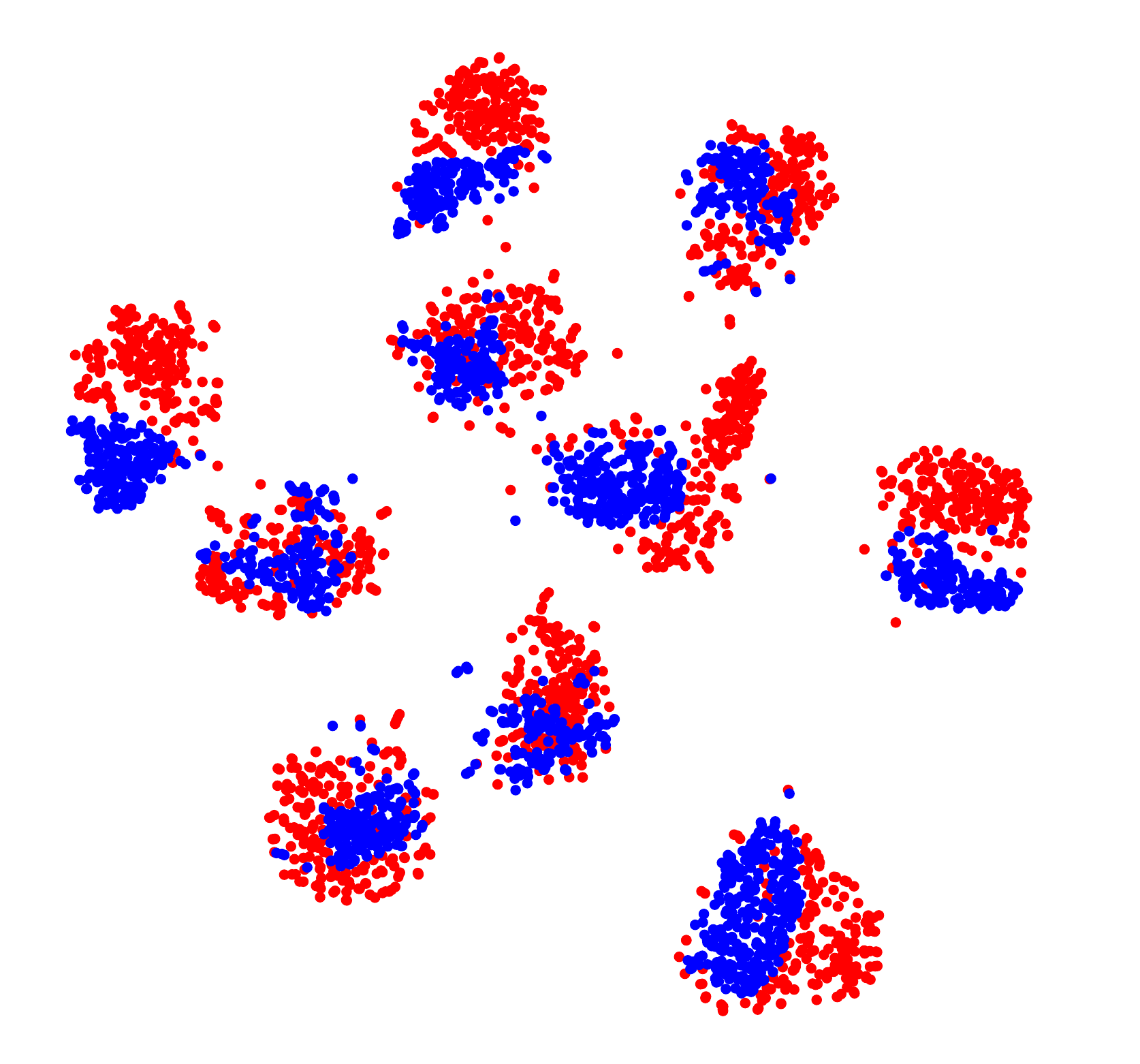} \\
(a) Epoch: 0 &(b) Epoch: 5&(c) Epoch: 25 &(d) Epoch: 50 &(e) Epoch: 150\\
%\includegraphics[width=4cm]{pictures/our_SourceOnlyCM(p-i)tensor(81)_2.pdf}&
%\includegraphics[width=4cm]{pictures/our_CM(p-i)tensor(93)_60.pdf}&
%\includegraphics[width=4cm]{pictures/our_TargetOnlyCM(p-i)tensor(99)_7.pdf}\\
%$(d) Source-only (p$\rightarrow$i)&(e) DAML (p$\rightarrow$i)&(f) Train-on-target (p$\rightarrow$i)\\
\end{tabular}
\caption{The t-SNE visualizations of features generated by the proposed DWL with the increse of the epoch on MNIST $\rightarrow$ USPS. Red and blue points indicate the source and target samples, resp.}
\label{t-SNE}
\end{figure*}
\subsection{Experimental Analysis}
\textbf{Visualization of feature space. }Fig. \ref{t-SNE} depicts the t-SNE visualizations of features learned in our method on the MNIST to USPS. With the increase of training epochs from (a) to (e), we can observe that feature distribution at epoch 150 has shown good clustering effect.
The learned features align the source and target domain samples with 10 clusters with clear boundaries.

\begin{figure} [h]
\centering
\begin{tabular}{cc}
\includegraphics[width=3.9cm]{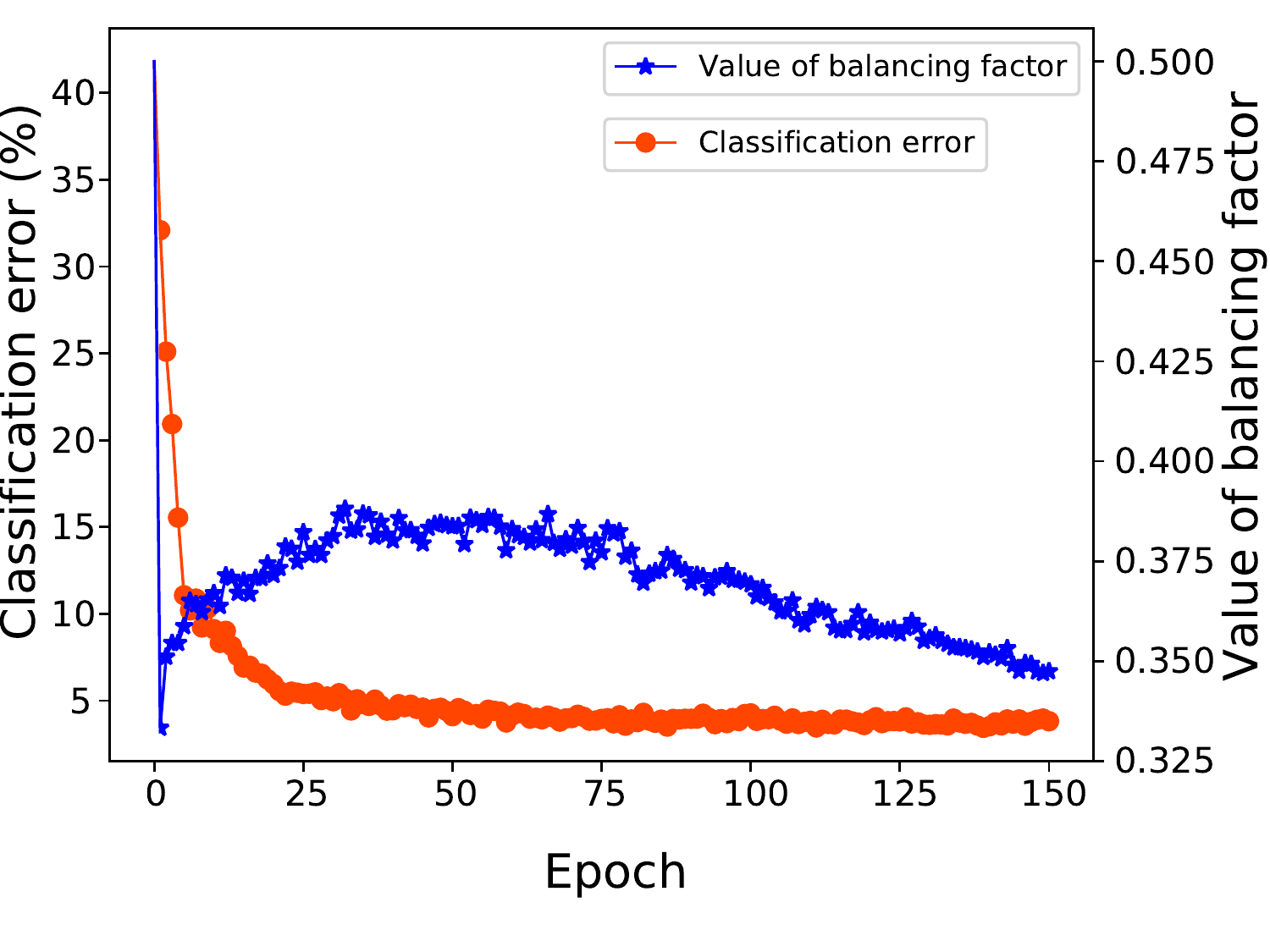}&
\includegraphics[width=4.1cm]{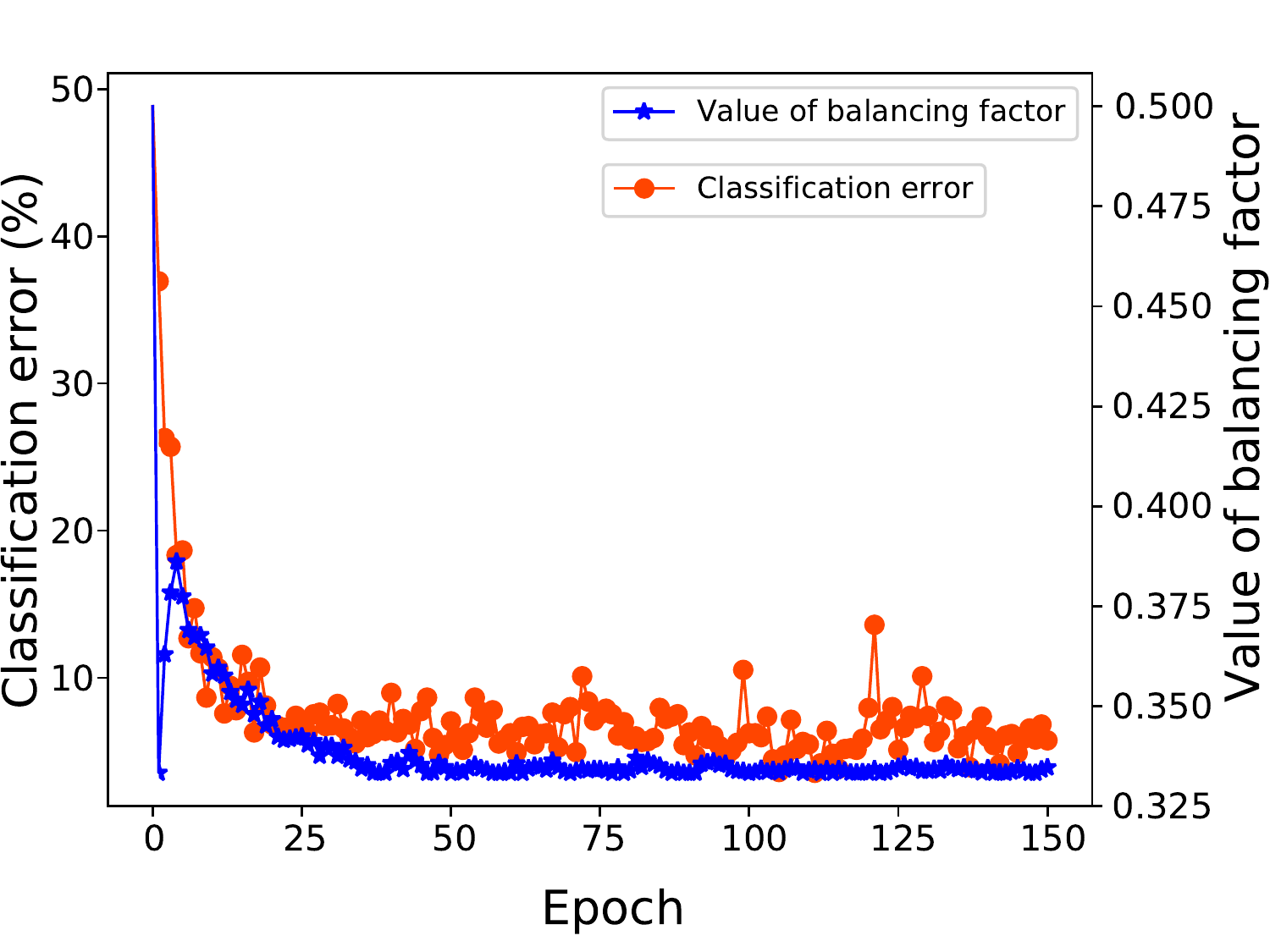}\\
%(a) USPS$\rightarrow$MNIST \\
%\includegraphics[width=7cm]{picture/u_m_converge.pdf}\\
%(b)  DSLR$\rightarrow$Webcam\\
\end{tabular}
\caption{Convergence analysis of DWL on the classification error (\%) and the dynamic balancing factor $\uptau$. }
\label{converge}
\end{figure}

\textbf{Convergence analysis.} The classification error and the value of balance factor $\uptau$ in our DWL on task USPS$\rightarrow$MNIST (left) and MNIST$\rightarrow$USPS (right) are shown in Fig. \ref{converge}. For each subfigure, the left axis of the red curve represents the classification error and the right axis of the blue curve represents the value of balancing factor $\uptau$. It can be found that both of them converge to a flat value gradually with iteration. This means that with the decrease of $\uptau$, the class discriminability is emphasised such that the classification error decreases too. In the iteration process, when the change of $\uptau$ is relatively obvious, the improvement of recognition accuracy is also relatively obvious. We set the initial value of $\uptau$ as 0.5, and it can be found that $\uptau$ drops sharply to less than 0.5 in the first epoch, indicating that the model has relatively good alignment but relatively poor discriminability.  Indeed, we know that the cross-domain digit dataset has small distribution difference but relatively poor discriminability, and therefore the results comply with an empirical observation.

\begin{figure*} [htb]
\centering
\begin{tabular}{cccc}
\includegraphics[width=4cm]{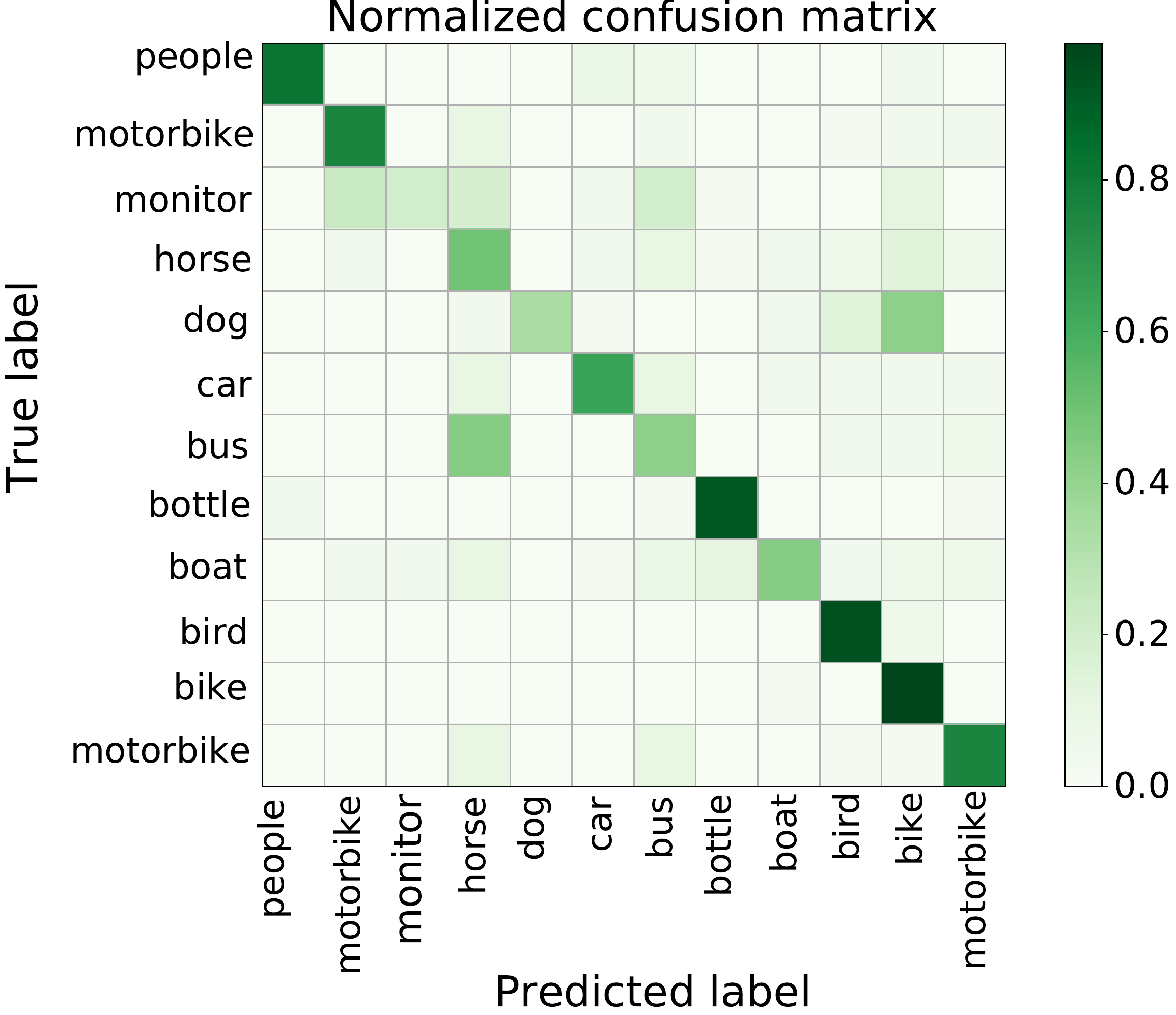}&
\includegraphics[width=4.1cm]{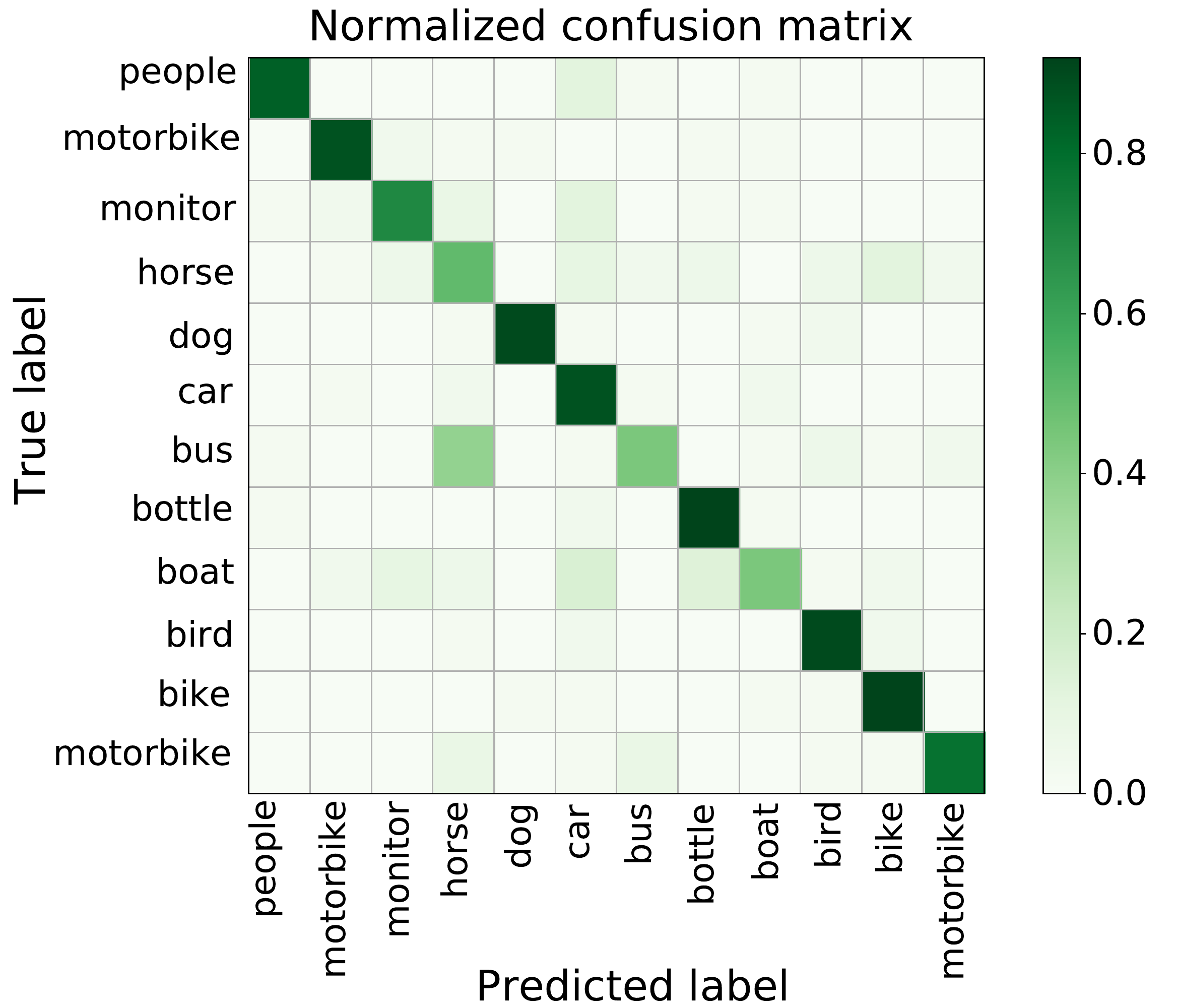}&
\includegraphics[width=4cm]{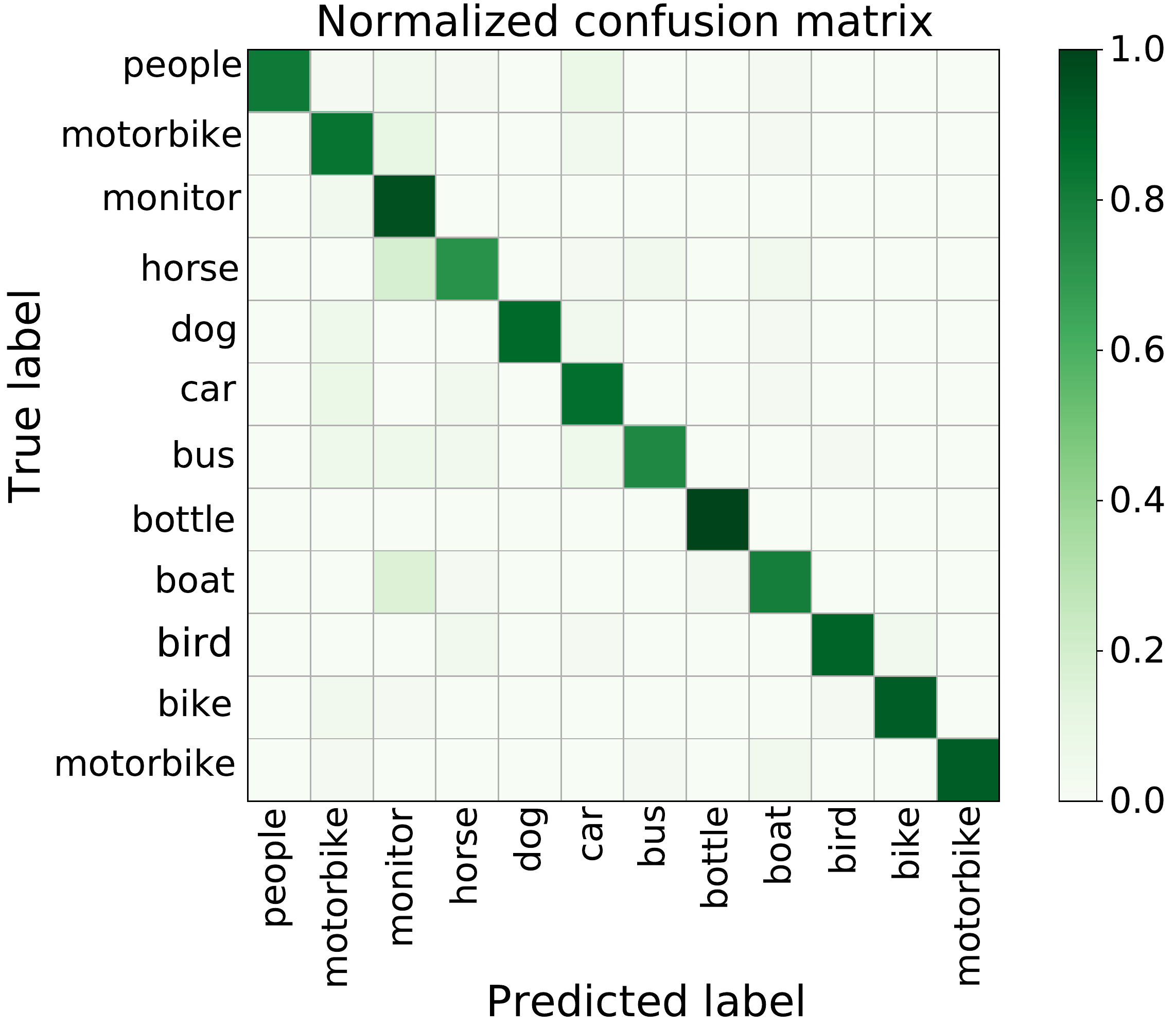}&
\includegraphics[width=4cm]{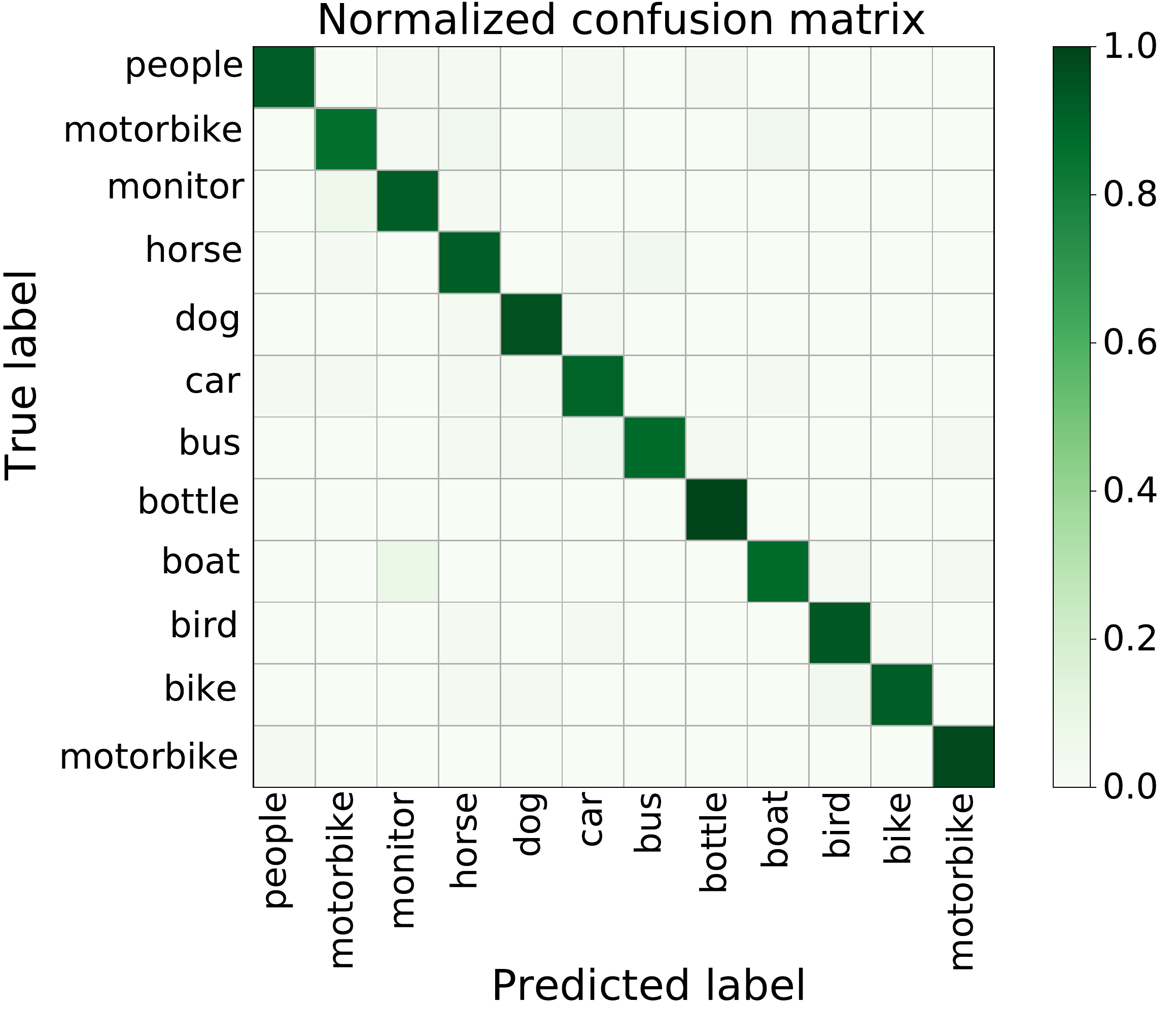}\\
(a) Source-only (c$\rightarrow$p)&(b) DWL (c$\rightarrow$p)& (c) Source-only (p$\rightarrow$i)&(d) DWL (p$\rightarrow$i)\\
\end{tabular}
\caption{The Confusion Matrix (CM) visualization for proposed DWL. }
\label{CM}
\end{figure*}

\textbf{Confusion matrix visualization.} Fig. \ref{CM} (a) to (d) show the visualizations of confusion matrix for the classifier trained by Source-only and our DWL on tasks c $\rightarrow$ p and p $\rightarrow$ i from ImageCLEF-DA. The Source-only means the classifier is trained only on labeled source data. Hence, the experimental results of Source-only can reveal the impact of domain discrepancy. As can be seen in Fig. \ref{CM}, the confusion is reduced for most classes through domain adaptation by our method. Comparing with Fig. \ref{CM} (a) and (b), the dog is misclassified as the bike for the Source-only and corrected in DWL. In p $\rightarrow$ i, the confusing horse is correctly recognized in DWL. The confusion matries further reveal the discriminability of our method on the target domain.
\begin{figure} [t]
\centering
\begin{tabular}{c}
\includegraphics[width=5cm]{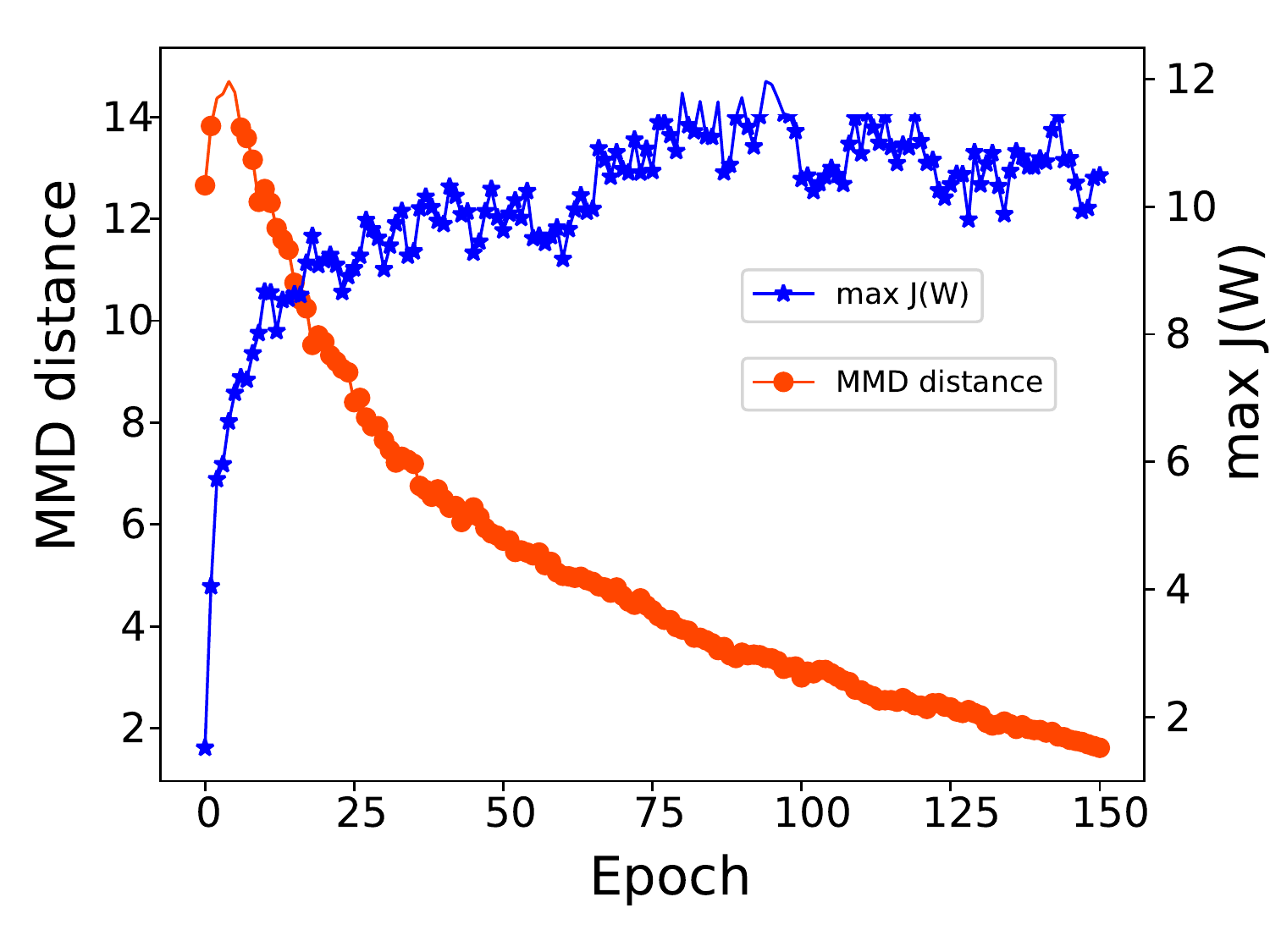}\\ %height=4.6cm,
%(a) USPS$\rightarrow$MNIST \\
%(b)  DSLR$\rightarrow$Webcam\\
\end{tabular}
\caption{Effectiveness analysis of DWL on alignment learning and discrimination learning. }
\label{effective}
\end{figure}
\textbf{Analysis of alignment and discriminability degrees.} The MMD distance and the max $J(\mathbf{W})$ of the feature representation learned in our DWL on task USPS$\rightarrow$MNIST are shown in Fig. \ref{effective}. It can be found that by dynamically weighting the alignment learning and discrimination learning, we make a better balance between them, which makes the two kinds of learning in an effective learning state and both of them can promote each other towards a good direction (i.e., the domain distance is smaller and the discriminability degree is larger).

\textbf{Ablation study.} In this part, we analyze the influence of sample weighting and balancing factor $\uptau$ on improving performance. The results are shown in Table \ref{tab_ablation1} and Table \ref{tab_ablation2}.
To analyze the role of sample weighting, the cross-domain task W$\rightarrow$A with large differences across domains in sample size is chosen for the experiment (The domain W contains 795 samples, and domain A contains 2817 samples). The experimental results are shown in Table \ref{tab_ablation1}. It can be found that the accuracy of the model can be significantly improved by introducing the sample weighting and dynamic balancing factor $\uptau$. Furthermore, we test the results of static weighting mechanism for domain alignment learning $\mathcal{L}_{da}$ and class discrimination learning $\mathcal{L}_{cd}$ on Digits dataset. As shown in Table \ref{tab_ablation2}, it can be found that the proposed dynamic weighting factor $\uptau$ achieves more excellent performance than the static weighting with different ratios. Our original intention for promoting the dynamic balance between domain alignment and class discrimination is confirmed.
\begin{table}[t]
%\scriptsize
\caption{The accuracy (\%) of ablation experiment.}
\begin{center}
\setlength{\tabcolsep}{2.1mm}{
\begin{tabular}{cc|c}
\hline
 Sample weighting & Balancing  factor $\uptau$ &Accuracy (\%) \\
\hline
x&x &67.9  \\
$\surd$ & x &68.3  \\
x& $\surd$ &69.1  \\
$\surd$ & $\surd$ &\bf{69.8} \\
\hline
\end{tabular}}
\end{center}
\label{tab_ablation1}
\end{table}
\begin{table}[!tb]
%\scriptsize
\caption{Accuracy (\%) of different weight between $\mathcal{L}_{da}$ and $\mathcal{L}_{cd}$}
\begin{center}
\setlength{\tabcolsep}{2.1mm}{
\begin{tabular}{c|ccccc|c}
\hline
Task & 1:1  & 1:9  & 9:1 &  7:3 & 3:7 &  $\uptau$ : $1-\uptau$ \\
\hline
M$\rightarrow$U &93.8 &96.2  &94.6  &94.5  &94.9  &\bf{97.3} \\
U$\rightarrow$M &96.8 &95.6  &96.8  &95.6  &95.6  &\bf{97.4} \\%
%D$\rightarrow$W & &  &  &  &  & \\
%W$\rightarrow$D & &  &  &  &  & \\
%P$\rightarrow$I & &  &  &  &  & \\
%I$\rightarrow$P & &  &  &  &  & \\
\hline
\end{tabular}}
\end{center}
\label{tab_ablation2}
\end{table}
\section{Conclusions}
In this paper, we analyze the interaction between alignment learning and discrimination learning and then propose that the weight of alignment learning and discrimination learning should be better dynamically balanced. For a more generalized model to accommodate different datasets, we propose the Dynamic Weighted Learning (DWL) method to dynamically adjust the weights of domain alignment learning and class discrimination learning to adapt to different cross-domain classification scenarios. Besides, the problem of model bias during training which is caused by the imbalanced sample size of the two domains is also considered in this paper, and we propose a simple but effective sample weighting mechanism to solve it.
Extensive experiments demonstrate that the dynamic learning idea between the $\mathcal{H}\Delta\mathcal{H}$-divergence and the combined error $\lambda$ is useful and scalable in DA.
{\small
\bibliographystyle{ieee_fullname}
\bibliography{dwl_arXiv}
}
\end{document}